\documentclass[runningheads]{llncs}
\usepackage[T1]{fontenc}
\usepackage{graphicx,verbatim}
\usepackage{amsmath}
\usepackage{amsfonts}
\usepackage{booktabs} 
\usepackage{multirow}
\begin{document}
\title{Self-supervised Normality Learning and Divergence Vector-guided Model Merging for Zero-shot Congenital Heart Disease Detection in Fetal Ultrasound Videos}
%
\titlerunning{Zero-shot Congenital Heart Disease Detection in Fetal Ultrasound Videos}
\authorrunning{Saha et al.}

\author{Pramit Saha\inst{1}\textsuperscript{*} \and 
        Divyanshu Mishra\inst{1}\textsuperscript{*} \and 
        Netzahualcoyotl Hernandez-Cruz\inst{1} \and 
        Olga Patey\inst{2} \and 
        Aris Papageorghiou\inst{2} \and 
        Yuki M. Asano\inst{3} \and 
        J. Alison Noble\inst{1}
}

\institute{Department of Engineering Science, University of Oxford
\and Nuffield Department of Women's and Reproductive Health, University of Oxford
\and Fundamental AI Lab, University of Technology Nuremberg}

\renewcommand{\thefootnote}{\fnsymbol{footnote}}
\footnotetext[1]{Indicates equal contribution}

    
\maketitle              

\begin{abstract}
 
Congenital Heart Disease (CHD) is one of the leading causes of fetal mortality, yet the scarcity of labeled CHD data and strict privacy regulations surrounding fetal ultrasound (US) imaging present significant challenges for the development of deep learning-based models for CHD detection. Centralised collection of large real-world datasets for rare conditions, such as CHD, from large populations requires significant co-ordination and resource. In addition, data governance rules increasingly prevent data sharing between sites.  To address these challenges, we introduce, for the first time, a novel privacy-preserving, zero-shot CHD detection framework that formulates CHD detection as a normality modeling problem integrated with model merging. In our framework dubbed Sparse Tube Ultrasound Distillation (STUD), each hospital site first trains a sparse video tube-based self-supervised video anomaly detection (VAD) model on normal fetal heart US clips with self-distillation loss. This enables site-specific models to independently learn the distribution of healthy cases.
To aggregate knowledge across the decentralized models while maintaining privacy, we propose a Divergence Vector-Guided Model Merging approach, DivMerge, that combines site-specific models into a single VAD model without data exchange. Our approach preserves domain-agnostic rich spatio-temporal representations, ensuring generalization to unseen CHD cases. We evaluated our approach on real-world fetal US data collected from 5 hospital sites. Our merged model outperformed site-specific models by 23.77\% and 30.13\% in accuracy and F1-score respectively on external test sets.


\end{abstract}

\section{Introduction and Background}

Congenital heart disease (CHD) accounts for approximately 28\% of all congenital anomalies worldwide \cite{van2011birth}. CHD encompasses a diverse range of heart conditions, varying in frequency and severity, and can be diagnosed early by fetal ultrasound (US) scanning. As a widely used non-invasive screening tool, fetal US is favored for its rapid data acquisition, affordability, portability, and ability to perform assessments without ionizing radiation. Early detection of CHD from fetal US is crucial to ensure long-term health outcomes \cite{khalil2013fetal,carvalho2002improving,becker2006detailed}. However, diagnosing CHD remains challenging and time-intensive due to the subtle nature of certain heart defects and variable fetal US video quality. Further, fetal heart assessment~\cite{van2009methods} presents significant challenges due to several factors, including fetal movement, rapid heart rate, small size, and limited accessibility. This suggests a clinical need for approaches to automated CHD detection, and an opportunity for deep learning-based analysis. However, applying supervised deep learning methods is often impractical, as they are not designed for highly imbalanced data scenarios such as our clinical setting.   Many forms of CHD are extremely rare, resulting in highly imbalanced datasets. Conversely, a large volume of fetal ultrasound videos from healthy fetuses is routinely collected during standard screening procedures. In this work, we exploit the availability of healthy population video to train a novel anomaly detection framework to identify CHD cases during inference.

\begin{figure}[t]
    \centering
\includegraphics[width=1\columnwidth]{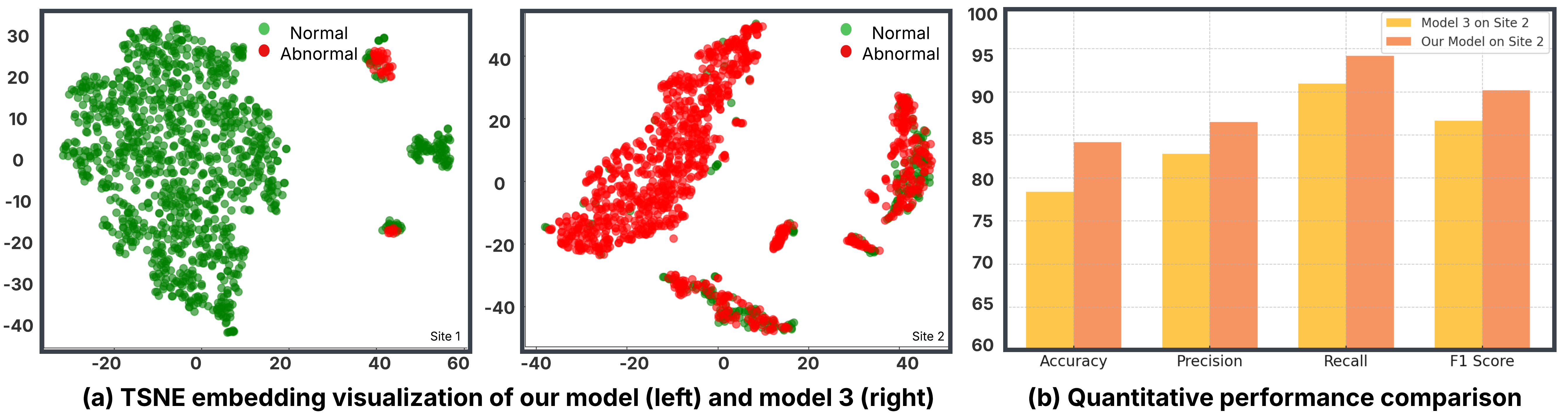}
    \caption{(a) t-SNE visualization (left) shows that our proposed method (after merging models trained on 3 sites) achieves nearly distinct clustering, suggesting well-separated feature representations. On the other hand, Model 3 (trained on Site 3) is observed to achieve low separability (right). (b) The quantitative comparison of both models evaluated on Site 2 further illustrates the benefit of our proposed model merging technique.} 
\label{fig:1}
\end{figure}

Anomaly detection models in the literature, such as \cite{yu2022deep,park2023normality,zhao2022lgn,ahn2024videopatchcore}, offer promising solutions for normality modeling. For effective performance, such models need to be trained on diverse, centralized data sourced from multiple hospital sites, thereby capturing a broad spectrum of fetal cardiac variations in appearance, geometry and disease. However, privacy regulations prohibit cross-hospital data sharing \cite{shojaei2024security}, creating a significant bottleneck in development of centralized models. To this end, instead of combining data from different sites, we propose to effectively merge models trained at individual sites. Our approach allows the aggregation of knowledge learned from individual hospital datasets without the need for data sharing and while avoiding interference and conflicts due to domain shifts. This ensures that essential task information is preserved, leading to a merged model that leverages the strengths of each local model while maintaining privacy compliance. 
The primary contributions of this work are as follows:

\noindent
1. To the best of our knowledge, this is the first work to introduce \textbf{video normality learning for CHD detection in fetal US videos}. We train a self-supervised video anatomy detection network on healthy fetal US clips using self-distillation loss that incorporates a student-teacher model \textbf{(\S 2.1)}. Our novel \textbf{Sparse Tube Ultrasound Distillation (STUD)} model learns the spatio-temporal representation related to healthy fetal hearts and is leveraged to detect previously unseen CHD anomalies during test time. Our model is light-weight as we sparsely sample 3D space-time tubes of varying sizes from the US video to create learnable tokens, which are then processed by a vision transformer. This enables us to develop strong and computationally efficient video models.

\noindent
2. This is also the first work to investigate \textbf{model merging for multi-site US analysis as well as for normality modeling}. We propose a two-step model merging procedure dubbed DiVMerge to enhance robustness with respect to model noise and model drifts while preserving normality information \textbf{(\S 2.2)}.  We first compute the geometric median of local models, acting as a denoising mechanism and then compute the divergence vectors as the difference between individual models and the geometric median. The parameters with small divergence vector component are retained, while others are replaced by the geometric median. In addition, the overall magnitude of the divergence vector for each site model is used to dynamically weight the updated local models before merging. 

\noindent
3. Our method enables \textbf{zero-shot CHD detection} coupling of the normality model and k-Nearest Neighbours (KNN) algorithm, thereby eliminating the need for additional fine-tuning. Trained on healthy data from 3 hospitals in a privacy-preserving manner, our normality model demonstrates the ability to detect anomalies \cite{carvalho2023isuog} (such as Hypoplastic Left Heart Syndrome (HLHS), Coarctation of the Aorta (COA), Right Aortic Arch (RAA), Left Superior Vena Cava (LSVC), Ventricular Septal Defects (VSD), and Cardiomegaly (CM)). In particular, DiVMerge outperforms the centralized model and all individual site-specific models on 2 external hospital datasets with distinct domain shifts.

\section{Methodology}
\subsection{Site-specific Self-supervised Video Anomaly Detection}

Ultrasound videos are inherently fine-grained and require dense temporal sampling to capture subtle changes essential for accurate understanding and detection. However, conventional tokenization methods using 2D patching \cite{jiang2024timeformer} or fixed 3D kernels \cite{tong2022videomae} generate an excessive number of tokens, making dense sampling computationally expensive and reducing the number of frames that can be processed on limited computing. To address this, our Sparse Tube Ultrasound Distillation (STUD) network employs a sparse tube sampling \cite{piergiovanni2023rethinking} that drastically reduces token redundancy while preserving spatio-temporal detail. For self-supervised video normality modeling, we integrate a video-focused self-distillation loss inspired by DINO \cite{caron2021emerging}, which trains a teacher-student network to learn consistent feature representations across diverse augmented views.
\\
\textbf{Sparse Tube Construction and Feature Extraction}
We adopt a sparse sampling strategy to address the limitations of dense tokenization. A standard 2D convolution with a $16 \times 16$ kernel is applied on frames sampled with a large temporal stride (e.g., every 16th frame). The total number of tokens generated from a video clip of dimensions $T \times H \times W$ is defined by: $N_{\text{tokens}} = \frac{T}{s_T} \times \frac{H}{s_H} \times \frac{W}{s_W}$ where $s_T$, $s_H$, and $s_W$ denote the temporal and spatial strides, respectively.
\begin{figure}[t]
    \centering
\includegraphics[width=1.0\columnwidth]{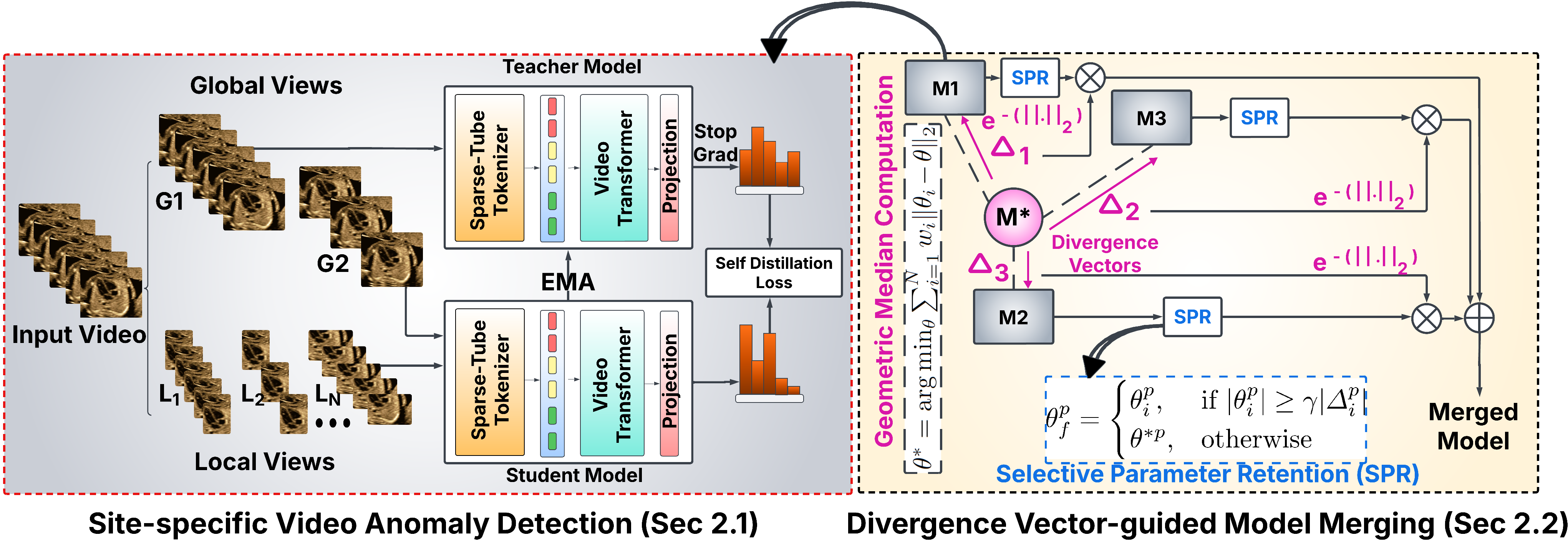}
    \caption{Overview of the proposed technique. Left figure shows self-supervised video anomaly network training at each site leveraging sparse-tube tokenizer and teacher-student model via self-distillation loss. This leads to the development of models $M_1$, $M_2$, and $M_3$ at three sites. The right figure shows the geometric median computation followed by estimation of divergence vectors for each site. The divergence vectors are then employed for selective parameter retention to reduce model drift and for adaptively weighting different models for final model merging.} 
\label{fig:1}
\end{figure}
We adapt two primary types of tubes: \textbf{(a) Image Tubes} of shape $1 \times 16 \times 16 \times d$ (where $d$ is hidden dimension), 
which tokenize individual frames and \textbf{(b) Video Tubes} of shape $8 \times 8 \times 8 \times d$, which capture the spatio-temporal context over multiple frames. Both types of tubes use a stride of $16 \times 16 \times 16$. 
To further capture diverse motion patterns in ultrasound videos, we incorporate variations such as temporally elongated tubes ($16 \times 4 \times 4$) for long-duration actions and spatially focused tubes ($2 \times 16 \times 16$) for fine spatial detail (see Fig. 2).
A space-to-depth transformation is applied reduce the channel dimension of the feature map (by a factor of 2), effectively enlarging the receptive field without increasing the number of parameters. Additionally, we learn a single 3D kernel ($8 \times 8 \times 8$) reshaped by trilinear interpolation to adapt to various tube configurations ($4 \times 16 \times 16$ or $32 \times 4 \times 4$). These enhancements ensure that our sparse sampling method captures all the necessary details while reducing computational demands. 

\noindent
\textbf{Self-Supervised Learning via Self-Distillation in the Video Domain} We generate multiple spatio-temporal augmentations (each called a 'view') to capture global context and fine-grained local details. Specifically, we create two global views encompassing a large portion of the video's temporal and spatial dimensions and eight local views focusing on smaller, more detailed regions (see Fig. 2). In our self-distillation framework, the teacher network processes only the global views to produce target feature representations, while the student network processes both global and local views.
The self-distillation loss encourages the student representations to align with the teacher by minimizing the discrepancy between their outputs. To ensure stable learning, the teacher model parameters are updated by an exponential moving average (EMA) of the student model.

\subsection{Divergence Vector-guided Model Merging (DiVMerge)}

 \begin{figure}[t]
    \centering
\includegraphics[width=1\columnwidth]{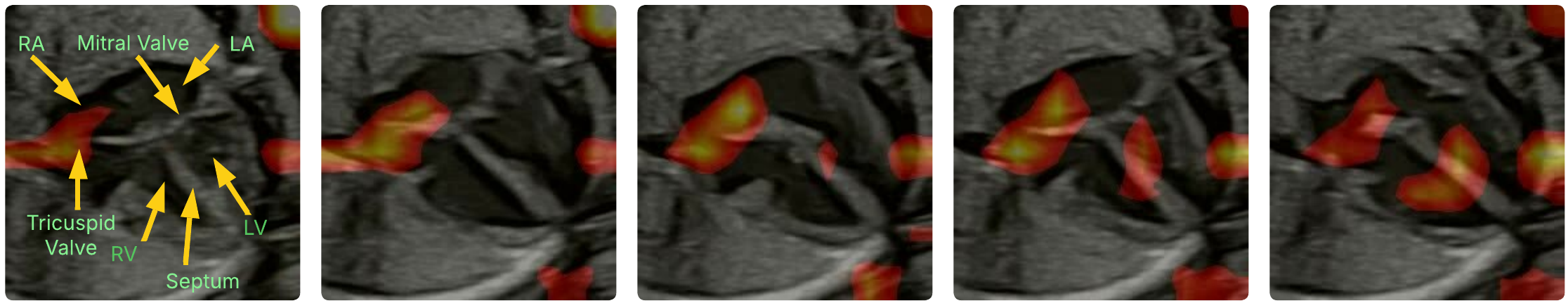}
    \caption{Feature map visualization overlaid on sequential US frames, highlighting the model's capability to focus on key anatomical fetal heart structures for CHD} 
\label{fig:1}
\end{figure}

We propose a two-step model merging procedure designed to improve robustness to model drifts and interference while preserving essential normality information. 
\\
\subsubsection{Geometric Median Computation:} The first step involves computing the geometric median of the locally trained models from individual sites (see Fig. 2). 
Given a set of \( N \) locally trained models \( \mathcal{M} = \{ M_1, M_2, \dots, M_N \} \), each represented by its parameter set \( \theta_i \in \mathbb{R}^d \), the geometric median is the point \( \theta^* \) that minimizes the sum of Euclidean distances to all individual models:
$\theta^* = \arg\min_{\theta} \sum_{i=1}^{N} w_i \|\theta_i - \theta\|_2$ where \( \theta_i \) represents the parameter vector of the \( i \)-th local model, \( w_i \) is the weighting factor for each model (uniform in our case). This method is particularly beneficial in our distributed settings, where model updates may vary significantly across clients due to the natural heterogeneity in data distributions.  The geometric median filters out inconsistencies, outliers, and extreme deviations in model updates that may arise due to small or biased datasets, noisy labels, or domain shifts in individual hospitals.
\\
\subsubsection{Divergence Vector-Based Adaptation:}
We define the divergence vector for each site model as the difference between the local model trained on that site and the geometric median model. 
For each model \( M_i \), the divergence vector \( \Delta_i \) is computed as: $\Delta_i = \theta_i - \theta^*$ where \( \theta_i \) represents the parameter vector of the locally trained model at site \( i \), \( \theta^* \) is the geometric median computed in Step 1, and \( \Delta_i \) represents the site-specific deviation from the geometric median. This vector captures how much each model deviates from the robust median representation and serves two key purposes:  \textbf{(A) Dynamic Model Weighting:} It adaptively assigns importance to the model contributions of each site in the final weighted averaging step.  A model with a small divergence vector is likely to be more stable and reliable, whereas a model with a large divergence vector may reflect domain-specific biases or noise. We use the magnitude of the divergence vector to assign dynamic weights $\alpha_i$ to each site model, allowing models to contribute proportionally based on their distance from the geometric median where $\alpha_i = \exp(-\lambda \|\Delta_i\|_2)$. 
\( \lambda \) is a scaling factor that controls the influence of the magnitude of the divergence vector, \( \|.\|_2 \) is the L2-norm. The final normalized weight for model \( i \) is: $\tilde{\alpha}_i = \frac{\alpha_i}{\sum_{j=1}^{N} \alpha_j}$. The final merged model \( \theta_f \) is computed as:
$\theta_f = \sum_{i=1}^{N} \tilde{\alpha}_i \theta_i$. 
This adaptive weighting reduces the effect of noisy local models.
\\
 \textbf{(B) Selective Parameter Retention:} The divergence vector is also utilized to localize normality information at the parameter level. Each parameter $\theta^p$ is retained only if its weight magnitude exceeds a $\gamma-$rescaled magnitude of the divergence vector, as it likely represents a confident and stable feature. Otherwise, it is replaced by the corresponding geometric median parameter to prevent excessive deviation from the global consensus. $\theta_f^p =
\begin{cases} 
    \theta_i^p, & \text{if } |\theta_i^p| \geq \gamma |\Delta_i^p| \\
    \theta^{*p}, & \text{otherwise}.
\end{cases} $

\section{Experiments and Results}

\noindent
\subsection{Dataset and Experimental Settings}
The five datasets used in this work are fetal heart ultrasound video sweeps collected from five hospitals: John Radcliffe Hospital (Oxford, UK) (Site 1), St George’s University Hospital (London, UK) (Site 2), Royal Brompton Hospital (London, UK) (Site 3),  Gold Coast Hospital (Queensland, Australia) (Site 4) and Chelsea \& Westminster Hospital (London, UK) (Site 5). Data were acquired using 10 different ultrasound machines by sonographers and fetal cardiologists. Healthy fetal US videos from Sites 1 to 3 (train / test: 8,878/667, 16,074/2,088, and 1,573/667, respectively) were used to train the model. The anomaly evaluation set from these hospitals included cases of COA and HLHS. The test data from external Sites 4 and 5 (29 and 18 samples, respectively) were only reserved for zero-shot testing and comprised cases of four other anomalies, \textit{viz.},  RAA, LSVC, VSD, and CM. All videos (mean length: 125 frames) were pre-processed with a cropping model to extract the heart region while removing patient and site-specific details.

\noindent
\subsection{Training and Implementation Details}
During training, we randomly sampled a clip of 64 frames per video with a sampling rate of 3, while during evaluation we uniformly sampled N clips from each video to extract features. A KNN classifier was then applied to these clip features, classifying a video as an anomaly if any clip was flagged abnormal. For self-distillation, we used 2 global crops (size 224) and 8 local crops (size 96), applying spatial transforms: color jittering, solarization, Gaussian blur and varying temporal sampling rates for both crop types. All models were trained for 200 epochs on a RTX6000 GPU (VRAM 25GB) with a batch size of 12 and a cosine learning rate schedule with a 5e-04 initial learning rate. 
Scaling factor $\lambda$ was fixed at 0.005 via grid search. 
\subsection{Performance analysis of site-specific Video Anomaly Detection}
Table 1 shows the comparison of our model (trained and tested individually at each site) with two baseline methods \textit{viz.}, TimeFormer \cite{jiang2024timeformer} (w/ supervised pre-training) and VideoMAE \cite{tong2022videomae} (w/ self-supervised pre-training) in detecting normal and abnormal fetal heart clips. Our model provides a favorable trade-off between computational efficiency and predictive performance. TimeFormer achieves slightly higher accuracy for some sites at the cost of 10x more tokens. VideoMAE underperforms compared to our method and TimeFormer for all internal sites. The highest overall F1-score of our model shows its best generalization capability and most stable performance across different sites while reducing computational costs.
Figure 3 shows a visualization of the attention maps from the final layer.  This demonstrates effective localization of our model on sequential US video frames around the tricuspid valve, foramen ovale flap, and inter-ventricular septum which are key anatomical structures for CHD detection. 

\begin{table}[htbp]
    \centering
    \renewcommand{\arraystretch}{1.2}
\caption{Performance of site-specific models for CHD detection in sites 1-3}
\scalebox{1.0}{
    \begin{tabular}{l|c|cc|cc|cc|cc}
        \toprule
        \multirow{2}{*}{Model} & \multirow{2}{*}{\textbf{\# tokens}} & \multicolumn{2}{c|}{\textbf{Site 1}} & \multicolumn{2}{c|}{\textbf{Site 2}} & \multicolumn{2}{c|}{\textbf{Site 3}} & \multicolumn{2}{c}{\textbf{Average}} \\
        \cline{3-10}
        & & \textbf{Prec} & \textbf{F1} & \textbf{Prec} & \textbf{F1} & \textbf{Prec} & \textbf{F1} & \textbf{Prec} & \textbf{F1} \\
        \midrule
        TimeFormer  & 12522 (100\%) & 40.00 & 57.14 & 81.13 & 88.61 & \textbf{89.37} & \textbf{93.85} & 70.17 & 79.87 \\
        VideoMAE & 6272 (50\%) & 43.48 & 46.51 & 69.95 & 78.62 & 88.13 & 93.09 & 67.19 & 72.74 \\
        Ours & \textbf{1176 (9.39\%)} & \textbf{48.72} & \textbf{64.41} & \textbf{87.13} & \textbf{90.97} & 83.84 & 90.14 & \textbf{73.66} & \textbf{82.20} \\ 
        \bottomrule
    \end{tabular}}
\end{table}

\subsection{Performance analysis of Model Merging}
\subsubsection{Evaluation on sites 1-3} Table 2 shows the performance (precision and F-1 score) of the model built using DiVMerge on sites 1, 2, and 3, compared to the centralized model (\textit{i.e.}, trained on data combined from all sites) and individual models (\textit{i.e.} models trained locally on their own sites). We also compare with 5 SOTA model merging methods, \textit{viz}., Model Soup (2022)\cite{wortsman2022model}, Task vector (2023) \cite{ilharcoediting}, Ties Merging (2023) \cite{yadav2023ties}, DARE (2024) \cite{yu2024language}, Model Stock (2024) \cite{jang2024model}. Note that all SOTA methods other than Model Soup need a base model for task vector computation whereas our merging strategy does not require one. This enhances the usability and potential scope of application of our model, including for scenarios where a base model is unavailable.
We observe that divergence-guided merging of models (ours) is almost as good as or better than the model built from centralized data. In addition, the overall performance of our model is higher than that of individual local models. While performance slightly drops with respect to individual models for Site 2 (which has the highest amount of data), our model outperforms individual models trained on Sites 1 and 3, which have less data. The improvement in precision is 3.83\% for Site 3 and 4.06\% for Site 1. This reveals a benefit of model merging for sites with limited data availability. In addition, this shows that our merged model eliminates the need to store site-specific models. While Model Soup and other SOTA merging techniques have a slight performance drop due to model drifts induced by heterogeneous sites, our model shows overall stable performance. This demonstrates that the use of the divergence vector can effectively mitigate the inter-site model conflicts.

\begin{table}[htbp]
    \centering
    \renewcommand{\arraystretch}{1.2}
\caption{Performance comparison of model merging for sites 1-3. (B.M = Base Model)}
\scalebox{1.0}{
 \begin{tabular}{l|c|cc|cc|cc|cc}
        \toprule
      \multirow{2}{*}{Model} & \multirow{2}{*}{\shortstack{\textbf{B.M.} \\ \textbf{needed?}}} & \multicolumn{2}{c|}{\textbf{Site 1}} & \multicolumn{2}{c|}{\textbf{Site 2}} & \multicolumn{2}{c|}{\textbf{Site 3}} & \multicolumn{2}{c}{\textbf{Average}} \\
        \cline{3-10}
        & & \textbf{Prec} & \textbf{F1} & \textbf{Prec} & \textbf{F1} & \textbf{Prec} & \textbf{F1} & \textbf{Prec} & \textbf{F1} \\
        \midrule
        Centralized     & - & 50.00  & 65.52  & 87.30  & 91.20  & 89.45  & 93.98  & 75.58  & 83.57  \\
        Individual & - & 48.72  & 64.41  & 87.13  & 90.97  & 83.84  & 90.14  & 73.23  & 81.84  \\ \midrule
        Model Soup \cite{wortsman2022model}    &  No & 41.30  & 57.58  & \textbf{86.75}  & 89.79  & {86.34}  & 92.45  & 71.74  & 79.94  \\
        Task Vector \cite{ilharcoediting}    & Yes  & 35.19     & 51.35      & {76.86}	& 88.53      & 88.52      & 93.54     & 66.85	  & 77.00     \\
        Ties Merging \cite{yadav2023ties}  & Yes  & 32.60      & 44.44      & 79.45     & 85.98      & {89.42}  & 93.50      & 67.16      &  74.64     \\
        DARE \cite{yu2024language}      & Yes  & 32.60	   & 44.40      & 83.00     & 86.80      & {88.40}  & 93.60      & 68.00      & 74.93     \\
        Model Stock \cite{jang2024model}  & Yes  & 36.70  & 52.20      & 84.27    & 88.11     & \textbf{89.68}     & \textbf{94.18}      & 70.21     &  78.16 \\
      
        \textbf{Ours ($\gamma=0.01$)}  & No & \textbf{52.78}  & \textbf{67.86}  & {86.50}  & \textbf{90.18}  & {87.67}  & {92.31}  & {75.65}  & {83.45}  \\

        \textbf{Ours ($\gamma=0.1$) } & No & \textbf{52.78}  & \textbf{67.86}  & {86.21}  & {89.64}  & {87.67}  & \textbf{{94.18}}  & \textbf{76.22}  & \textbf{83.89}  \\

     
        \bottomrule
    \end{tabular}}
    \label{tab:performance_comparison}
\end{table}

\subsubsection{Evaluation on sites 4 and 5}
 Sites 4 and 5 are different geographical locations and cover different patient demographics to those used to train the merged model. Evaluation of our merge model on  sites 4 and 5 data is shown in Tab. 3 and Fig. 4. These show that our model generalises well to these new data scenarios where there are domain shifts due to different ultrasound scanners and data acquisition procedures at these sites. The centralized model and Model 1 fails to detect most abnormal cases resulting in an overall F1 score of 23.55 and 20.0 respectively. By contrast, Model 3 considers almost all abnormal cases as healthy hearts. Model 2 performs better but misidentifies 15 out of 20 healthy samples as CHD. Our model achieves the best performance, improving on the centralized model by 20\% and 54.6\% in accuracy and F1 score respectively. 

\begin{figure}[t]

    \centering
\includegraphics[width=1\columnwidth]{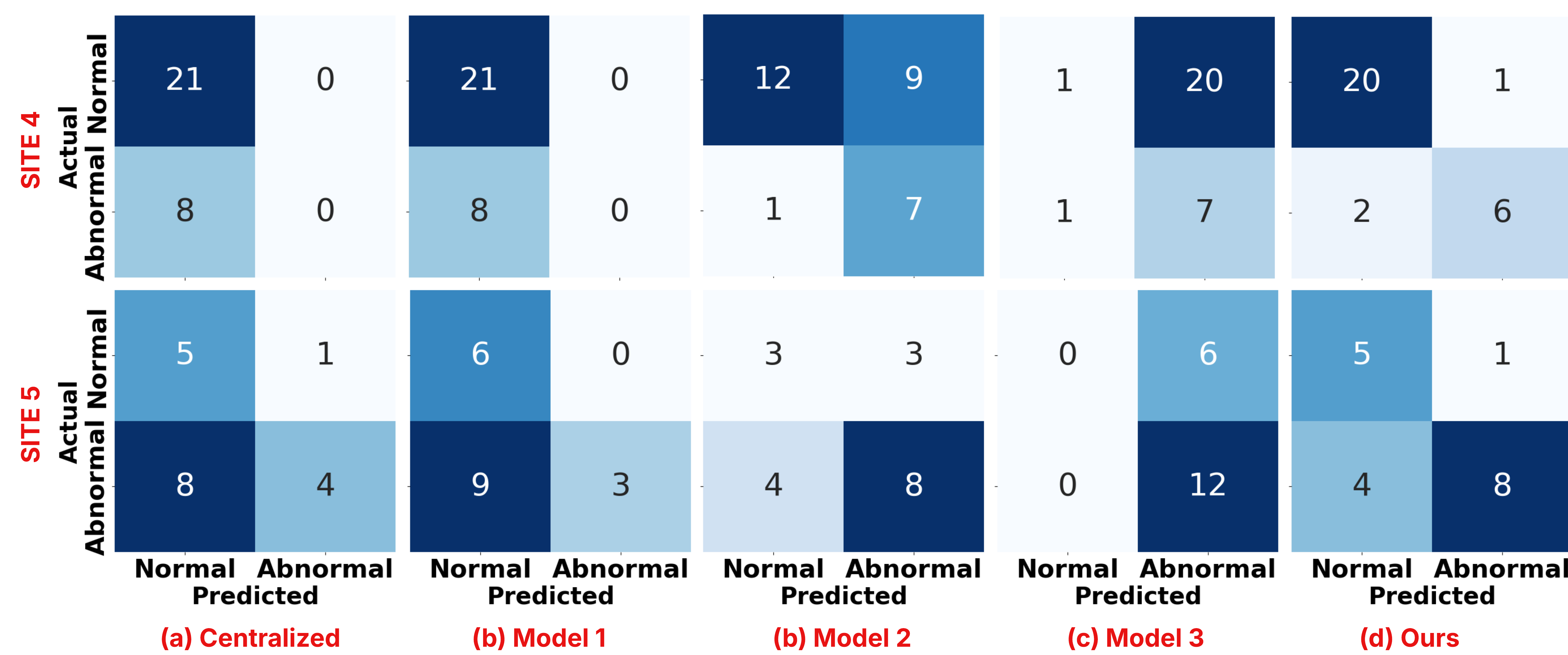}
    \caption{Confusion matrices illustrating the performance of various models on external sites 4 and 5. The results indicate that the Centralized Model and Model 1 struggle to detect most abnormal cases, while Model 3 frequently misclassifies normal cases as abnormal due to domain shift. In contrast, our model achieves the best performance, accurately distinguishing most normal and abnormal cases even with domain gap.} 
\label{fig:1}
\end{figure}

\begin{table}[htbp]
    \centering
    \renewcommand{\arraystretch}{1.2}
    \caption{Performance comparison of merged model w.r.t. baselines on Sites 4 and 5}
    \scalebox{0.8}{
    \begin{tabular}{l|cccc|cccc|cccc}
        \toprule
        \multirow{2}{*}{Model} & \multicolumn{4}{c|}{\textbf{Site 4}} & \multicolumn{4}{c|}{\textbf{Site 5}} & \multicolumn{4}{c}{\textbf{Average}} \\
        \cline{2-13}
        & \textbf{Accuracy} & \textbf{Prec} & \textbf{Recall} & \textbf{F1} 
        & \textbf{Accuracy} & \textbf{Prec} & \textbf{Recall} & \textbf{F1} 
        & \textbf{Accuracy} & \textbf{Prec} & \textbf{Recall} & \textbf{F1} \\
        \midrule
        Centralized  & 72.0  & 0.0   & 0.0   & 0.0   & 50.0  & 80.0  & 33.3  & 47.1  & 61.0  & 40.0  & 16.7  & 23.5  \\
        Model 1       & 72.4  & 0.0   & 0.0   & 0.0   & 50.0  & 100.0 & 25.0   & 40.0  & 61.2  & 50.0  & 12.5   & 20.0   \\
        Model 2       & 65.5  & 43.8  & 87.5  & 58.3  & 61.1  & 72.7  & 66.7   & 69.6  & 63.3  & 58.3  & 77.1   & 63.9  \\
        Model 3       & 27.6  & 25.9  & 87.5  & 40.0  & 66.7  & 66.7  & 100.0  & 80.0  & 47.2  & 46.3  & 93.8  & 60.0   \\
        \textbf{Ours} & \textbf{89.7}  & \textbf{85.7}  & \textbf{75.0}  & \textbf{80.0}  
                      & \textbf{72.2}  & \textbf{88.9} & \textbf{66.7}   & \textbf{76.2}  
                      & \textbf{81.0} & \textbf{87.3}  & \textbf{70.9}   & \textbf{78.1}  \\
        \bottomrule
    \end{tabular}}
\end{table}

\section{Conclusion}
In this work, we have introduced a novel privacy-preserving, zero-shot CHD detection framework that leverages self-supervised video normality learning (STUD) and a novel divergence vector-guided model merging (DiVMerge) technique to overcome the challenges of scarce labeled data and cross-hospital privacy constraints in fetal US. 
Evaluation with real-world fetal US datasets from five hospitals demonstrates that our method achieves superior performance than SOTA methods in detecting CHD anomalies. 
Compared with a model built with centralized data, our merged model achieves a 20\% improvement in accuracy and a 54.6\% increase in F1-score on unseen test data sets, highlighting its strong zero-shot generalization capabilities even under domain shifts caused by variations in ultrasound scanners, acquisition procedures, and patient demographics.

Our results highlight that model merging can serve as a viable alternative to centralized learning in privacy-sensitive clinical settings. Our experiments show that model merging has the potential to improve performance particularly for sites with limited data through the knowledge acquired from models trained on other sites via merging. Another notable observation is that our merged model significantly outperforms model trained with centralised data by mitigating domain shifts, thanks to its adaptive weighting of site-specific models and the resolution of inter-site model conflicts via the proposed divergence vector.



%
\bibliographystyle{splncs04}
\bibliography{ref}

\begin{thebibliography}{10}
\providecommand{\url}[1]{\texttt{#1}}
\providecommand{\urlprefix}{URL }
\providecommand{\doi}[1]{https://doi.org/#1}

\bibitem{ahn2024videopatchcore}
Ahn, S., Jo, Y., Lee, K., Park, S.: Videopatchcore: An effective method to memorize normality for video anomaly detection. In: Proceedings of the Asian Conference on Computer Vision. pp. 2179--2195 (2024)

\bibitem{becker2006detailed}
Becker, R., Wegner, R.D.: Detailed screening for fetal anomalies and cardiac defects at the 11--13-week scan. Ultrasound in Obstetrics and Gynecology: The Official Journal of the International Society of Ultrasound in Obstetrics and Gynecology  \textbf{27}(6),  613--618 (2006)

\bibitem{caron2021emerging}
Caron, M., Touvron, H., Misra, I., J{\'e}gou, H., Mairal, J., Bojanowski, P., Joulin, A.: Emerging properties in self-supervised vision transformers. In: Proceedings of the IEEE/CVF international conference on computer vision. pp. 9650--9660 (2021)

\bibitem{carvalho2023isuog}
Carvalho, J., Axt-Fliedner, R., Chaoui, R., Copel, J., Cuneo, B., Goff, D., Gordin~Kopylov, L., Hecher, K., Lee, W., Moon-Grady, A., et~al.: Isuog practice guidelines (updated): fetal cardiac screening. Ultrasound Obstet Gynecol  \textbf{61}(6),  788--803 (2023)

\bibitem{carvalho2002improving}
Carvalho, J., Mavrides, E., Shinebourne, E., Campbell, S., Thilaganathan, B.: Improving the effectiveness of routine prenatal screening for major congenital heart defects. Heart  \textbf{88}(4),  387--391 (2002)

\bibitem{ilharcoediting}
Ilharco, G., Ribeiro, M.T., Wortsman, M., Schmidt, L., Hajishirzi, H., Farhadi, A.: Editing models with task arithmetic. In: The Eleventh International Conference on Learning Representations

\bibitem{jang2024model}
Jang, D.H., Yun, S., Han, D.: Model stock: All we need is just a few fine-tuned models. In: European Conference on Computer Vision. pp. 207--223. Springer (2024)

\bibitem{jiang2024timeformer}
Jiang, D., Ke, Z., Zhou, X., Hou, Z., Yang, X., Hu, W., Qiu, T., Guo, C.: Timeformer: Capturing temporal relationships of deformable 3d gaussians for robust reconstruction. arXiv preprint arXiv:2411.11941  (2024)

\bibitem{khalil2013fetal}
Khalil, A., Nicolaides, K.H.: Fetal heart defects: potential and pitfalls of first-trimester detection. In: Seminars in fetal and neonatal medicine. vol.~18, pp. 251--260. Elsevier (2013)

\bibitem{park2023normality}
Park, S., Kim, H., Kim, M., Kim, D., Sohn, K.: Normality guided multiple instance learning for weakly supervised video anomaly detection. In: Proceedings of the IEEE/CVF winter conference on applications of computer vision. pp. 2665--2674 (2023)

\bibitem{piergiovanni2023rethinking}
Piergiovanni, A., Kuo, W., Angelova, A.: Rethinking video vits: Sparse video tubes for joint image and video learning. In: Proceedings of the IEEE/CVF Conference on Computer Vision and Pattern Recognition. pp. 2214--2224 (2023)

\bibitem{shojaei2024security}
Shojaei, P., Vlahu-Gjorgievska, E., Chow, Y.W.: Security and privacy of technologies in health information systems: A systematic literature review. Computers  \textbf{13}(2), ~41 (2024)

\bibitem{tong2022videomae}
Tong, Z., Song, Y., Wang, J., Wang, L.: Videomae: Masked autoencoders are data-efficient learners for self-supervised video pre-training. Advances in neural information processing systems  \textbf{35},  10078--10093 (2022)

\bibitem{van2011birth}
Van Der~Linde, D., Konings, E.E., Slager, M.A., Witsenburg, M., Helbing, W.A., Takkenberg, J.J., Roos-Hesselink, J.W.: Birth prevalence of congenital heart disease worldwide: a systematic review and meta-analysis. Journal of the American College of Cardiology  \textbf{58}(21),  2241--2247 (2011)

\bibitem{van2009methods}
Van~Mieghem, T., DeKoninck, P., Steenhaut, P., Deprest, J.: Methods for prenatal assessment of fetal cardiac function. Prenatal Diagnosis: Published in Affiliation With the International Society for Prenatal Diagnosis  \textbf{29}(13),  1193--1203 (2009)

\bibitem{wortsman2022model}
Wortsman, M., Ilharco, G., Gadre, S.Y., Roelofs, R., Gontijo-Lopes, R., Morcos, A.S., Namkoong, H., Farhadi, A., Carmon, Y., Kornblith, S., et~al.: Model soups: averaging weights of multiple fine-tuned models improves accuracy without increasing inference time. In: International conference on machine learning. pp. 23965--23998. PMLR (2022)

\bibitem{yadav2023ties}
Yadav, P., Tam, D., Choshen, L., Raffel, C.A., Bansal, M.: Ties-merging: Resolving interference when merging models. Advances in Neural Information Processing Systems  \textbf{36},  7093--7115 (2023)

\bibitem{yu2022deep}
Yu, G., Wang, S., Cai, Z., Liu, X., Xu, C., Wu, C.: Deep anomaly discovery from unlabeled videos via normality advantage and self-paced refinement. In: Proceedings of the IEEE/CVF Conference on computer vision and pattern recognition. pp. 13987--13998 (2022)

\bibitem{yu2024language}
Yu, L., Yu, B., Yu, H., Huang, F., Li, Y.: Language models are super mario: Absorbing abilities from homologous models as a free lunch. In: Forty-first International Conference on Machine Learning (2024)

\bibitem{zhao2022lgn}
Zhao, M., Zeng, X., Liu, Y., Liu, J., Li, D., Hu, X., Pang, C.: Lgn-net: local-global normality network for video anomaly detection. arXiv preprint arXiv:2211.07454  (2022)

\end{thebibliography}

\end{document}